\documentclass[a4paper,twoside]{article}
\usepackage{cite}
\usepackage{subcaption}
\usepackage{amsmath,amssymb,amsfonts}
\usepackage{algorithm}
\usepackage{algorithmicx}
\usepackage{algpseudocode}
\usepackage{graphicx}
\usepackage{textcomp}
\usepackage{xcolor}
\usepackage{url} 
\usepackage{amsthm}
\usepackage{multicol}
\usepackage{pslatex}
\usepackage{SCITEPRESS}     
\begin{document} 

\title{Towards computer vision technologies: \\Semi-automated reading of  automated utility meters 
}

\author{\authorname{Maria Spichkova, Johan van Zyl}
\affiliation{School of Computing Technologies, RMIT University}
\affiliation{Melbourne, Victoria 3000, Australia}
\email{maria.spichkova@rmit.edu.au}
}


\abstract{
In this report we analysed a possibility of using computer vision techniques
for automated reading of utility meters. In our study, we focused on two computer vision techniques: an open-source solution Tensorflow Object Detection (Tensorflow) and a commercial solution Anyline.  This report extends our previous publication \cite{spichkova2020application}: We start with presentation of  a structured analysis of related approaches.
After that we provide a detailed comparison of two computer vision technologies, 
Tensorflow Object Detection (Tensorflow) and Anyline, applied to semi-automated reading of utility meters. In this paper, we discuss limitations and benefits of each solution applied to utility meters reading, especially focusing on aspects such as accuracy and inference time. Our goal was to determine the solution that is the most suitable for this particular application area, where there are  several specific challenges.
}

\onecolumn \maketitle \normalsize \setcounter{footnote}{0} \vfill

\section{{Introduction}} 
\label{sec:intro}

\noindent 
Smart solutions are extensively used for many kind of utilities, e.g., electricity and gas.
Therefore, development and application of smart devices, including smart meters, is an active research topic over the last decades  \cite{depuru2011smart,benzi2011electricity,zheng2013smart}. 
Smart meters have many advantages, where the most important is the capability to record the data on energy consumption and sent this data automatically to the provider  and/or to the corresponding customer. 
Having this detailed data we can provide additional monitoring and billing features, as well as a much more detailed analysis of the consumption patterns. On this basis, we could also provide suggestions to a customer on how to reduce consumption or to schedule the energy-consuming tasks for the time, which is mostly suitable for the energy network (in the terms of payment or the energy load). 

However, even when smart utility meters provide many benefits and can be seen as the future direction of the industry, they also have a number of disadvantage that prevents an immediate replacement of old meters by the new smart solutions: 
\begin{itemize}
    \item 
    \emph{Cost:}
    Implementation of smart meters on a large scale is expensive. 
    In the case the upgrade is optional/voluntary,  customers might prefer to refuse or delay an upgrade to a smart meter, if they have to pay for an upgrade. In the case the upgrade is compulsory, customers might struggle with payment of an extensive bill, and a subsidy from government might be needed.
    \item
    \emph{Learning curve:}
    A customer can benefit from using a smart meter only if they know well how exactly to use it. 
    Some customers might refuse or delay an upgrade to a smart meter because they perceive that using smart meters is not so easy and requires reasonable learning efforts.
    \item
    \emph{Privacy concerns:}
    The data collected from the smart meters might be used to extract
    information on the usage pattern, which could provide a basis for identification whether/when the residents are currently at home, how many of residents are at home during the particular weekdays or day times, etc. This fact together with increased number of cyber-security attacks over recent years prevents some customers from using the smart meters.   
\end{itemize}

By these reasons, in some countries the roll out to the smart meter systems is done phase-wise, where on the initial phase the roll out is voluntary. As the initial phase might take many years, we believe that even during this period a semi-automated solution would be beneficial.  
For example, a semi-automated solution it might benefit  people with vision impairment. 

 \newpage 
 In this report we analysed a possibility of using computer vision techniques
for semi-automated reading of utility meters. This report extends our previous publication \cite{spichkova2020application}. We start with presentation of  a structured analysis of related approaches.
After that we provide a detailed comparison of two computer vision technologies, 
Tensorflow Object Detection (Tensorflow) and Anyline, applied to semi-automated reading of utility meters. We compare application of these technologies with each other, as well as
with our previous results presented in our earlier work \cite{enase2019shine}. 

In our earlier study,
we conducted a project in collaboration with Energy Australia, which is an electricity and gas retailing private company that supplies electricity and natural gas to more than 2.6 million residential and business customers throughout Australia. 
Their  solution for non-smart meters was to provide an online portal, where the consumers can update the records on the utility readings. While this solution provided many benefits for the customers during the initial phase of smart meters introduction, it also had some disadvantages: the consumers had to provide a lot of additional details, and to calculate their utility readings manually. 
The goal of our previous project was to elaborate an alternative method for the existing  
system, which would allow for a higher degree of automation to increase the usability of the system. 
The proposed solution was to use computer vision techniques for capturing readings. We analysed there the following computer-vision technologies: Google Cloud Vision (GCV), 
Amazon Web Services  (AWS)  Rekognition, 
Tesseract OCR, and Azure's Computer Vision.  The study demonstrated that AWS Rekognition provides better results for this application domain. However, it's accuracy was far from ideal: 
the average accuracy values AWS Rekognition was only $36\%$. 
In the current study, we used the same data sets, but applied other approaches. 
The current results  are significantly better  in the terms of recognition accuracy than the results of our early investigation study: 
the results of the current study demonstrate the accuracy up to $92.35\%$, which is promising for real-life application of the proposed solution.

\section{{Related Work and Background}}
\label{sec:related}

\subsection{Smart meters}

 The research on the smart meter devices and the corresponding analytic was actively conducted over many years, which was reflected not only in research publications but also in patents, see e.g., \cite{kelley2000automated,nap2001automatic,jenney1999automatic,knight1998remote,ehrke2003electronic,grady2016method,winter2017methods}. 
 
Over the last decade, 
there were two core research directions in this area: (1) privacy and security aspects of the smart meter application, and (2) smart meters in combination with a smart grid system. In the rest of the section we discuss the most cited (as per Google Scholar, retrieved 23 November 2022) 
publications, grouped by the research directions. 

A summary of most cited papers on {privacy and security aspects  of smart meters} is presented in Table~\ref{tab:privacy1}.  
    This research direction is currently the most active one among the mentioned directions, because the privacy and security concerns provide one of the biggest obstacles for the (potential) users of smart meters. 
    In many cases, data mining and data analytics techniques were applied on the meter reading data, to investigate the above issues questions.
A summary of most cited papers on design of smart meters for smart grid is presented in Table~\ref{tab:smartgrid2}.  

\begin{table*}[ht!]
\caption{Privacy and security aspects  of smart meters}
\label{tab:privacy1}
\centering 
\begin{tabular}{|l|l|l|}
\hline
Ref. & Description / core contributions & Citations \\  
\hline
\hline
\cite{molina2010private} & 
 Privacy-preserving smart meter architecture; a study
to demonstrate that 
& 605 \\
& the power consumption patterns can help to reveal how many people are in the home, & \\
& what are their sleeping and eating routines, etc.  &  ~\\     
\hline
\cite{greveler2012multimedia} & Multimedia content identification through smart meter power usage profiles & 256 \\
\hline
\cite{sankar2013smart} & Theoretical framework to analyse privacy aspects of smart meters  &  271 \\ 
\hline 
\cite{asghar2017smart} & Survey on smart meter data privacy & 203
\\
\hline
\cite{finster2015privacy} & Survey on privacy-aware smart metering & 97
\\
\hline
\cite{mckenna2012smart} & analysis on balancing consumer privacy concerns with legitimate applications & 406
\\
\hline
\cite{eibl2014influence} & Influence of data granularity on smart meter privacy & 112
\\
\hline
\cite{kleiminger2013occupancy} & An approach for occupancy detection from electricity consumption data & 268\\  
\hline
\cite{kleiminger2015household} & Household occupancy monitoring using electricity meters & 140\\
\hline
\cite{jin2017virtual} & Occupancy sensing using smart meters & 107\\
\hline
\cite{zou2017non} & Occupancy sensing using smart meters in commercial buildings & 96\\
\hline
\cite{rajagopalan2011smart} &    Formal framework to quantify the privacy trade off
problem in smart meter data  &   220 \\    
\hline
\cite{beckel2014revealing} & Extraction of the households characteristics from the the smart meter data   &   282 \\    
\hline
\cite{albert2013smart} & Analysis on customer profiles based the consumption patterns from the smart meter data   &  278    \\  
\hline
\cite{chen2013non} 
& Non-intrusive occupancy monitoring using smart meters,  energy-efficiency optimisations   &  207 \\ 
\hline
\cite{buchmann2013re}  & approach for re-identification of smart meter data &   82 \\  
\hline
\cite{bohli2010privacy} &
 A privacy model for smart metering & 265 \\
\hline    
\hline
\end{tabular}
\end{table*}

\begin{table*}[ht!]
\caption{Privacy and security aspects  of smart meters}
\label{tab:smartgrid2}
\centering 
\begin{tabular}{|l|l|l|}
\hline
Ref. & Description / core contributions & Citations \\  
\hline
\hline
\cite{efthymiou2010smart} & 
Approach for anonymizing  the data sent by a smart meter to a smart grid
& 784 \\  
\hline 
\cite{petrlic2010privacy} & 
A privacy-preserving concept for smart grids
& 90 \\ 
\hline 
\cite{li2010secure} & 
Smart meter data aggregation approach for smart grids
& 660 \\    
\hline 
\cite{zheng2013smart} & 
Overview of typical smart meter's aspects and functions  wrt. smart grid aspects
& 383 \\    
\hline 
\cite{xiao2013exploring} & 
Overview of approaches for energy theft recognition from the meter data
& 104 \\    
\hline 
\cite{arif6529714} & 
Study on design and development of smart energy meter for the smart grid
& 103 \\    
\hline 
\cite{avancini2019energy} & 
Analysis of energy meters evolution in smart grids  & 212\\
 \hline
\cite{yip2017detection}& Detection of energy theft and defective smart meters in smart grids using linear regression & 141 \\    
% ...
\hline    
\hline
\end{tabular}
\end{table*}



\subsection{Computer-vision technologies}

There are many  computer-vision technologies that are potentially applicable for semi-automated reading of smart meters, for example, 
\begin{itemize}
    \item Google Cloud Vision \footnote{\url{https://cloud.google.com/vision}} is a computer-vision technology provided by the Google platform ((Vertex AI Vision)),
    \item
    AWS  Rekognition\footnote{\url{https://aws.amazon.com/rekognition}} is another cloud-based solution, provided by Amazon Web Services,
    \item
    Tesseract OCR \cite{smith2007overview} is an  optical character recognition engine, which is released under the Apache License.
    \item
    Computer Vision AI service provided by Microsoft Azure\footnote{\url{https://azure.microsoft.com/en-gb/products/cognitive-services/computer-vision/}},
    \item
Tensorflow\footnote{\url{https://www.tensorflow.org/guide/summaries_and_tensorboard}} is an open-source machine learning system that operates at
large scale and offers a multitude of models to be retrained (more than 30), see \cite{abadi2016tensorflow,abadi2017computational}. It provides visualisation tool TensorBoard~\cite{wongsuphasawat2018visualizing}  that allows to visualise TensorFlow graphs, plot corresponding quantitative metrics, etc.
    \item
    Anyline\footnote{\url{https://anyline.com/products/ocr-meter-reading}} is a commercial solution intended to read utility meters, which also offers a free sample app \emph{Anyline OCR Scanner} that has been used during tests within our project. 
\end{itemize}

\subsection{Challenges of application of computer vision technologies}

There are many challenges for application of computer vision technologies for the reading of utility meters. In what follows we discuss the most critical of them:
\begin{itemize}
    \item 
    \emph{Reflections:} Most meter models encountered has a transparent protective cover, 
    which lead to reflections from it. This becomes problematic for computer vision technologies which includes thresholding/flooding techniques.  
    \emph{Thresholding techniques} are usually applied to OCR technologies in-order to minimize noise and or to convert to black-and-white images. \\
    \emph{Flooding} is a technique used to find similar neighbouring pixels. This technique can be used when finding contours of shapes. 
    
    \item
    \emph{Clipped digits:} The final digit in analogue meters usually rotate freely. This becomes problematic as the digit becomes clipped and the full digit is not displayed -- instead of this a half of one digit and a half of another digit are presented.  
    The computer vision technology would need to be able to deal with this issue and recognise clipped digits effectively.

\begin{figure*}[ht!] 
\begin{center}
\includegraphics[scale=0.85]{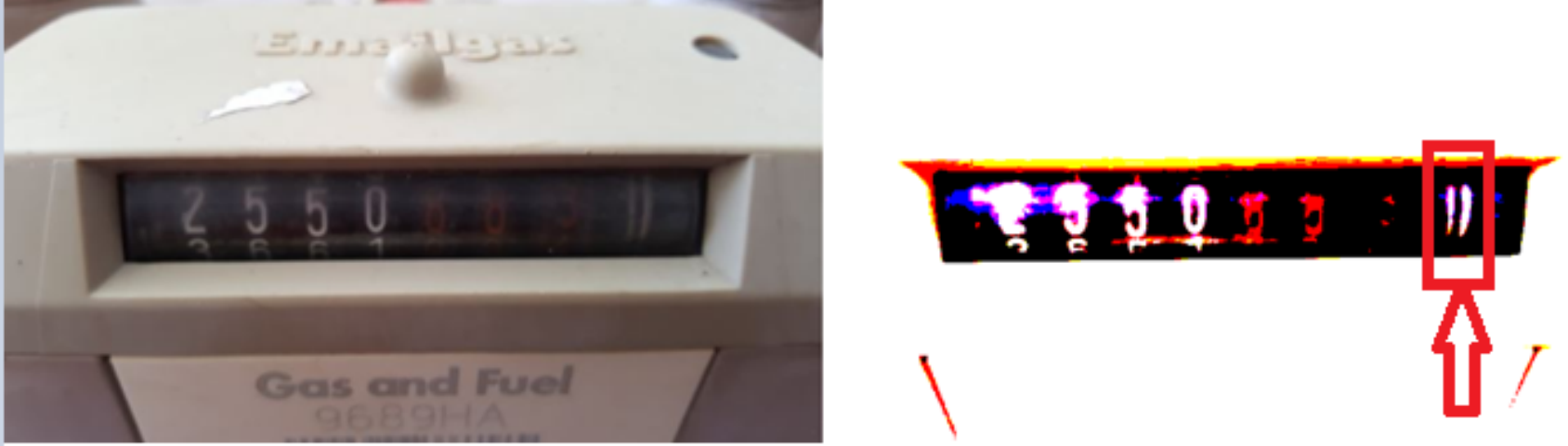} \\
(a)  \hspace{6cm} (b) 
\end{center}
\caption{Challenging case: (a) Original image; (b) Application of the thresholding technique over the original image\\ ~ \\ ~ \\ ~}
\label{fig:challenge}
\end{figure*}


    
     \item
    \emph{Not all characters and digits, which can be identified  on the meter, actually belong to the meter reading:} 
    Utility meters commonly include other text consisting of letters and digits. The computer vision technology would need to be able to discriminate which digits are part of the meter reading, and which have to be ignored.

   \item 
   \emph{Blur, noise, and warping:}
   Many meters observed during the project were not ideally clean as in real life the meters might become dirty and have grime even after few months of use. 
   The grime adds noise to the observed meter. Grime with reflection creates a blur effect around some digits. 
   Digits are also observed to be warped along with the shape of the cover, e.g., the digits appear ``stretched'' or ``squashed'' depending on observation angle.  Furthermore, there can be a significant contrast difference in scenarios where the colour of the digits is mixed. 

\begin{figure*}[ht!] 
\begin{center}
\includegraphics[scale=0.85]{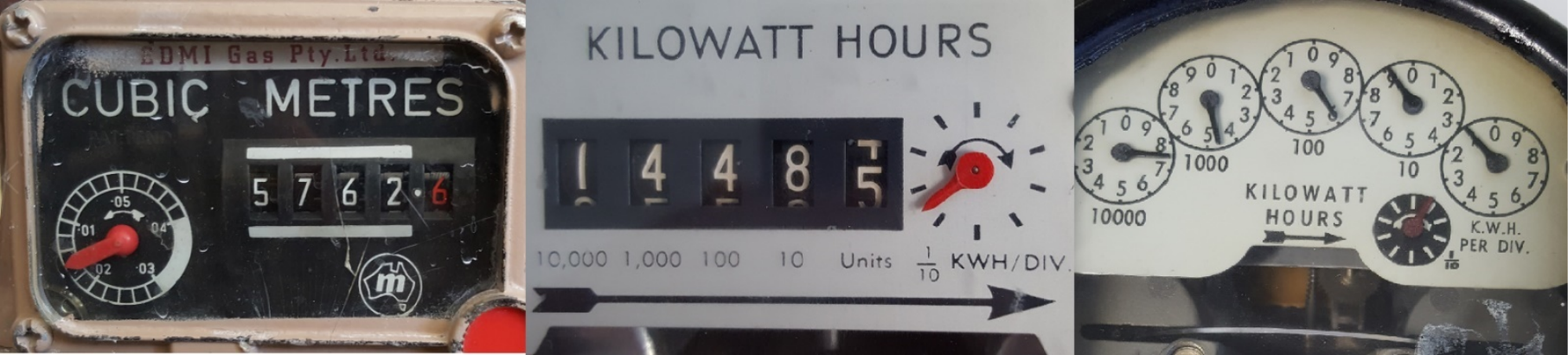} \\
(a) \hspace{4cm} (b) \hspace{4cm} (c)
\end{center}
\caption{Different representation styles (scales, dials and digit) mixed within a meter interface \\ ~\\ ~}
\label{fig:mixed}
\end{figure*} 

   \item
   \emph{
   Different representation styles (scales, dials and digits) mixed within a meter interface},  see Figure~\ref{fig:mixed}. 
   The reading value of utility meters is not represented in a consistent manner. 
   Utility models observed during the project presented their reading value either through: using rotational dials, using cyclometers, a combination of rotational dials and cyclometers or a singular digital display. 
The challenge becomes even more complex when the numeric value is displayed on a scale, e.g., as presented in Figure~\ref{fig:mixed}: 
\begin{itemize}
    \item 
    The meter (a)  should be read as 5762.615m$^3$ gas and not 57626,
	 \item  
	 The meter (b) should be read as 14485.68kWh energy and not 14485,
	 \item
	The meter (c) should be read as 75691.1kWh energy and not 756911.
\end{itemize}
The meters commonly include a decimal point. Most meters include the decimal point as the last digit and can be read and a tenth. Sometimes it is a mixture of a digit and a dial as in Figures~\ref{fig:mixed}(b) and (c). 

Some models display three decimal points as seen in Figure~\ref{fig:challenge} (the digits after decimal points are highlighted with red colour), which should read as 2550.883m$^3$ and not 2550883. It's easy to see from the image, that the red colour might make it more difficult to identify the digits in the case of manual reading. 
The aim of our work is to identify a computer vision technology that is capable of distinguishing between different numeric scales and be able to detect both digits and rotation of dials. 
\end{itemize}

\section{\uppercase{Methodology}}
\label{sec:methodology}

To determine which of these technologies is most suitable for reading utility meters, we elaborated a set of tests that allows us to identify the limitations of each technology by gradually adjusting image blur, noise, gamma or scale. 

The following methodology was applied to analyse the techniques:
\begin{enumerate}
    \item 
    To create a training dataset for the Tensorflow Object Detection framework. 
    \item 
 To train all the different models using the dataset, which was elaborated at Step 1. This was done through Google ML Engine. 
    \item 
 	To create evaluation datasets for which the technologies can be tested against. 
    \item 
 	To create a test harness for the involved technologies.
    \item 
 	To run the test harness on the evaluation datasets created from Step 3.
\end{enumerate}
The training dataset is a set of all images found during project duration.  The final training dataset consisted of 395 images and 2000 annotations. 
Unfortunately, this is still considered limited as supplied Tensorflow models were created based upon 2000+ annotations per object. 



As the evaluation dataset we used the same images as in our earlier work \cite{enase2019shine}, which allowed us to compare the results of application of Tensorflow and Anyline not only with each other, but also with the results of AWS  Rekognition (which demonstrated the best but not good enough accuracy in our previous study). 
Thus, in \cite{enase2019shine}, a total of 30 images were selected based on their ``uniqueness'' -- images with unique meters or images with unique lighting. These images were duplicated and modified with various effects in order to test the limitations of the different technologies. These effects are:
\begin{itemize}
    \item \emph{Scaling:} The dataset was scaled in steps of 0.1 ranging from a scale of 0.1 to 0.9 (10$\%$ to 90$\%$) of the original dataset. 
    \item \emph{Blurring:} Blurring was done in steps of 10 from 10 to 90 with an open source blur algorithm that is based on the normalised box filter, see \cite{OpenCV}. The algorithm uses a normalised box filter, the numeral value adjusts the kernel size. 
    %
    \item \emph{Gamma:} The gamma algorithm was used with an open source lookup table algorithm  \cite{OpenCV}.  The gamma correction to simulate different lightning conditions. 
    %
    \item \emph{Noise:} The noise algorithm is based upon the salt and pepper noise algorithm that adds sharp and sudden disturbances in the image in the form of sparsely occurring white and black pixels, see  \cite{Gonzalez:2001:DIP:559707}. This algorithm was included to further test the performance of the various technologies as noise arguably emulates ``dirt'' on meters. 
\end{itemize}

~\\
In contrast to Anyline, 
 Tensorflow's Tensorboard   provides more in-depth evaluation of each model and how well each model detects objects for a given dataset, such as: 
 \begin{itemize}
     \item  
     single-shot multi-box detector (SSD), see \cite{liu2016ssd},
     \item 
     feature pyramid networks (FPN), see \cite{lin2017feature},
     \item 
     fast region-based convolutional neural networks (FRCNN), see \cite{ren2017faster}.
 \end{itemize}

\noindent 
Our initial hypotheses in terms of accuracy and detection rates were as follows:
\begin{itemize}
    \item[H1:] FRCNN model would significantly outperform other models.
    \item[H2:] SSD model would perform significantly worse than the other models in terms of accuracy. 
    \item[H3:] The lower the image resolution is, the faster inference would occur. 
    \item[H4:] FRCNN would require significantly more time than FPN and SSD models.
\end{itemize}
%
As Tensorflow Object Detection only detects objects, the results have to be filtered in order to give a reading. 
We applied the filtering algorithm presented in Algorithm~\ref{alg:filtering} to
\begin{itemize}
    \item 
    remove any junk data, i.e., any detected objects for what the identification confidence is $\le 10\%$,
    \item
    remove all duplicates within the geometric region, keeping the results withe the highest identification confidence.
\end{itemize}

\begin{algorithm*}
\caption{Filtering of meter readings}
\label{alg:filtering}
\begin{algorithmic}[1] 
\State {j = 1}
\ForAll{$i \leftarrow 1, n$} 
\If {$confidence(result[i]) > 10$\% }  
\If {$result[i]$ is unique within geometric region} 
\State {$list[j] = result[i]$}
\State {$j  = j+1$}
\Else \Comment{where list[k] is the duplicate of result[i]}
    \If{confidence(result[i]) $>$ confidence(list[k])}
    \State {list[k] = result[i]}
    \EndIf
\EndIf
\EndIf
\EndFor
\State {Sort $list[i]$ by geometric location from left to right}
\end{algorithmic}
\end{algorithm*}

\noindent 
The filtering algorithm can be further improved, if the previous reading of the meter is considered, for which an access to the customer's account is required. 
We haven't applied this improvement within the comparison study, as the supplied Anyline app cannot include any customer related data. 

\noindent  
Results from each dataset and each technology were produced in 
csv files (a comma-separated values file that allows data to be saved in a tabular format) with the following structure:
\\

\texttt{
    FileName, InferTime, FilteredReading,}

\texttt{    ExpectedReading, IsCorrect
}\\

\noindent 
where \texttt{InferTime} denotes the interference time, which was measured in ms.
 
Similarly to \cite{enase2019shine}, we calculated the accuracy of recognition 
calculated as the following simple formula (we measure the accuracy in percents, where $100\%$ means a totally accurate recognition):
\begin{equation}
 Accuracy   = \frac{CorrectResults}{Total} *100
 \label{eq:e1}
\end{equation}
where \\
$CorrectResults$ is the number of results that  match with the original readings completely, \\
$Total$ presents the total number of images in a dataset. 
As in our study, we had 30 images in each of the datasets, $Total = 30$.

\section{\uppercase{Results and Discussion}}
\label{sec:results}

\begin{figure}[h!] 
\begin{center}
\includegraphics[scale=0.45]{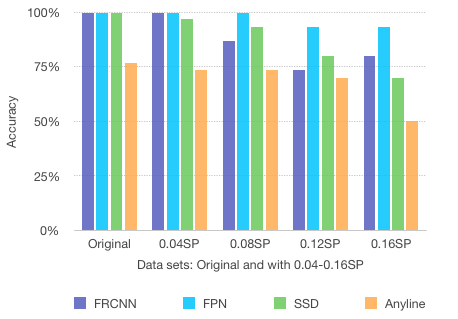}  
\end{center}
\caption{Accuracy scores per dataset:  Salt and pepper analysis}
\label{fig:Acc_sp}
\end{figure}

\begin{figure*}[ht!] 
\begin{center}
\includegraphics[scale=0.45]{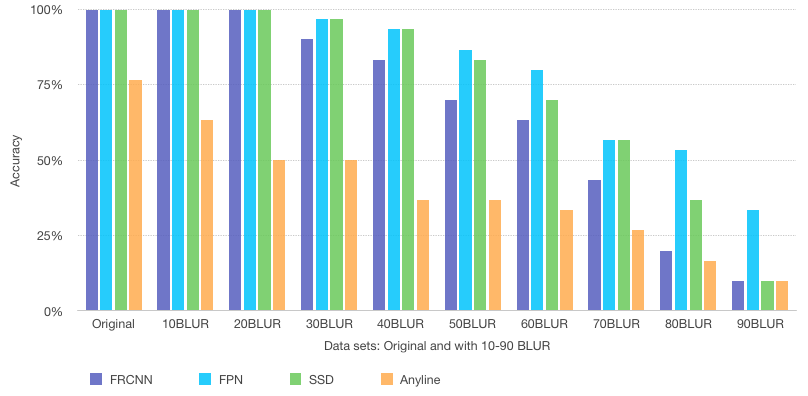}\\ 
\end{center}
\caption{Accuracy scores per dataset: Blur analysis }
\label{fig:Acc_blur}
\end{figure*}

\begin{figure*}[ht!] 
\begin{center}
\includegraphics[scale=0.45]{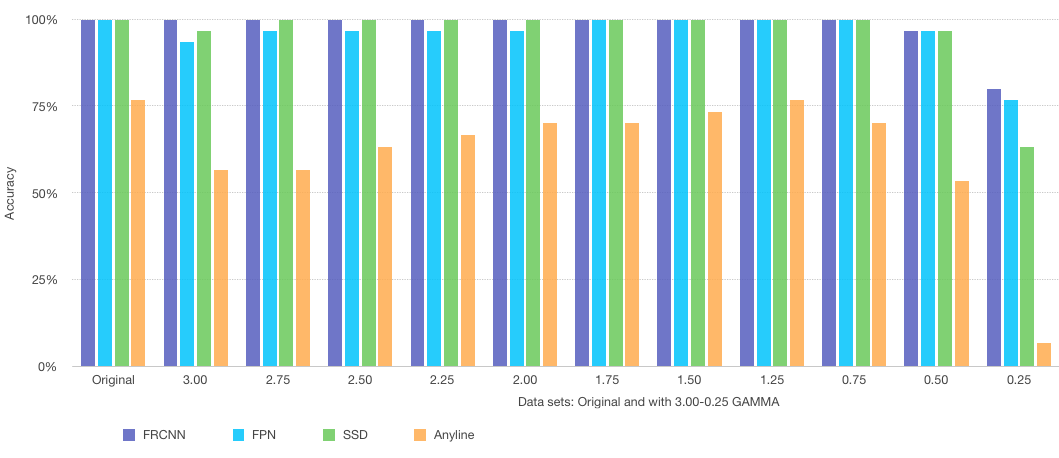} 
\end{center}
\caption{Accuracy scores per dataset: Gamma analysis }
\label{fig:Acc_gamma}
\end{figure*}

\begin{figure*}[ht!] 
\begin{center} 
\includegraphics[scale=0.5]{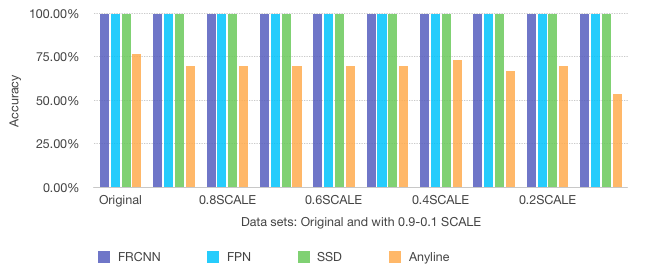} 
\end{center}
\caption{Accuracy scores per dataset: Scale analysis 
~\\~}
\label{fig:Acc_scale}
\end{figure*}

 Figures~\ref{fig:Acc_sp}-\ref{fig:Acc_scale}
 present the identified accuracy scores per dataset, where 50$\%$ indicates half of the dataset meter images were correctly read.

It is important to mention that both 
 Tensorflow  and Anyline were less sensitive to 
 salt and paper issues (Figure~\ref{fig:Acc_sp}), scaling (Figure~\ref{fig:Acc_scale}), gamma issues (Figure~\ref{fig:Acc_gamma}), and  than to blurring (Figure~\ref{fig:Acc_blur}). However, for all data sets, the accuracy of Tensorflow was almost twice higher:
 \begin{itemize}
     \item 
      With low blurring (10BLUR and 20BLUR), Tensorflow performed with 100$\%$ accuracy, where the accuracy of Anyline dropped to approx. 63$\%$ and 50$\%$ respectively. 
      \item 
      For the effect of 50BLUR, the accuracy of Tensorflow FPN and Anyline were  approx. 87$\%$ and 37$\%$ respectively.
      \item 
      For the effect of 90BLUR, the accuracy for Tensorflow FRCNN, Tensorflow SSD and Anyline was only 10$\%$,  
      where the accuracy of Tensorflow FPN was approx. 33$\%$. 
      However, the 90BLUR effect means a very blurry image.
 \end{itemize}

 \noindent 
The overall performance of the Tensorflow models greatly surpass expectations in terms of accuracy, having an average performance of $88.14\%$, $89.51\%$ and $92.35\%$ for FRCNN, SSD, and FPN, respectively.
 This is especially remarkable,  if we compare it with the average accuracy values from AWS Rekognition was only $36\%$ that demonstrated the best (but not really satisfactory) results within the study presented in \cite{enase2019shine}.  

   TensorBoard confirms the accuracy of the trained models. Scoring a near perfect score of 1.0 is extremely significant 
is a strong indication that Tensorflow Object Detection is a suitable framework for the automated meter reading. 
 
Anyline performed arguably well having an average performance of $57.16\%$, and struggled on several utility meter models. The results indicate that Anyline may not have trained or tested their product on a similar utility meters models as used within the Australian market. 

The accuracy of Anyline was significantly  lower for all data sets: if compared with Tensorflow FPN, the accuracy of Anyline was in average lower approx. $35\%$ lower, where 
\begin{itemize}
    \item 
    the largest differences in the cases of 0.25GAMMA (approx. $70\%$) and  20BLUR  (approx. $57\%$);
    \item
    the largest difference (approx. $23\%$) was in the cases of 1.25GAMMA, 0.12SP (noise), 90BLUR, and the original data sets.
\end{itemize}
~\\


  \begin{table}[h!]
    \centering
     \includegraphics[scale=0.67]{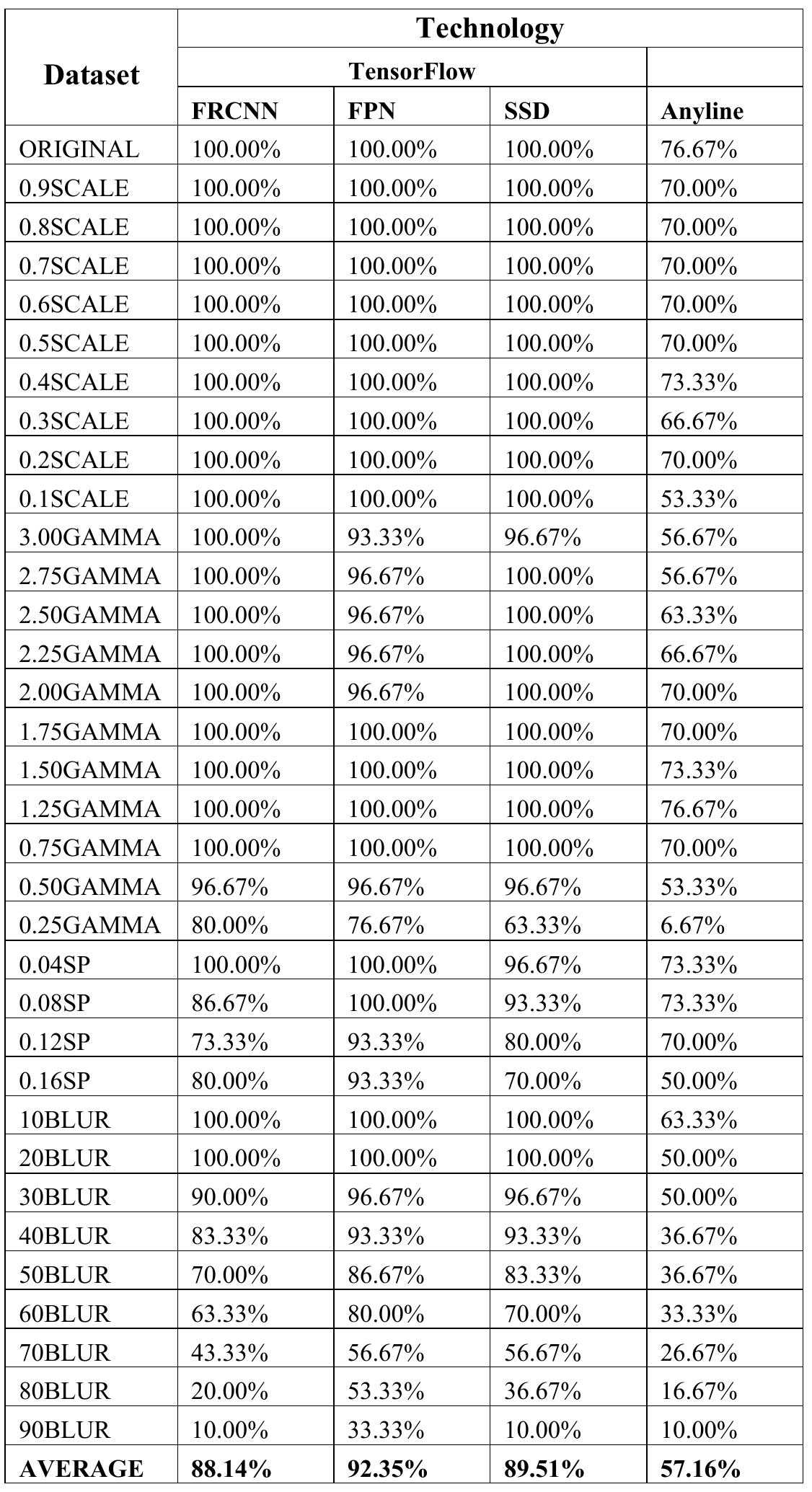}
    \caption{Accuracy scores per dataset }
    \label{table:acc1}
\end{table}

  \begin{table}[h!]
    \centering
     \includegraphics[scale=0.7]{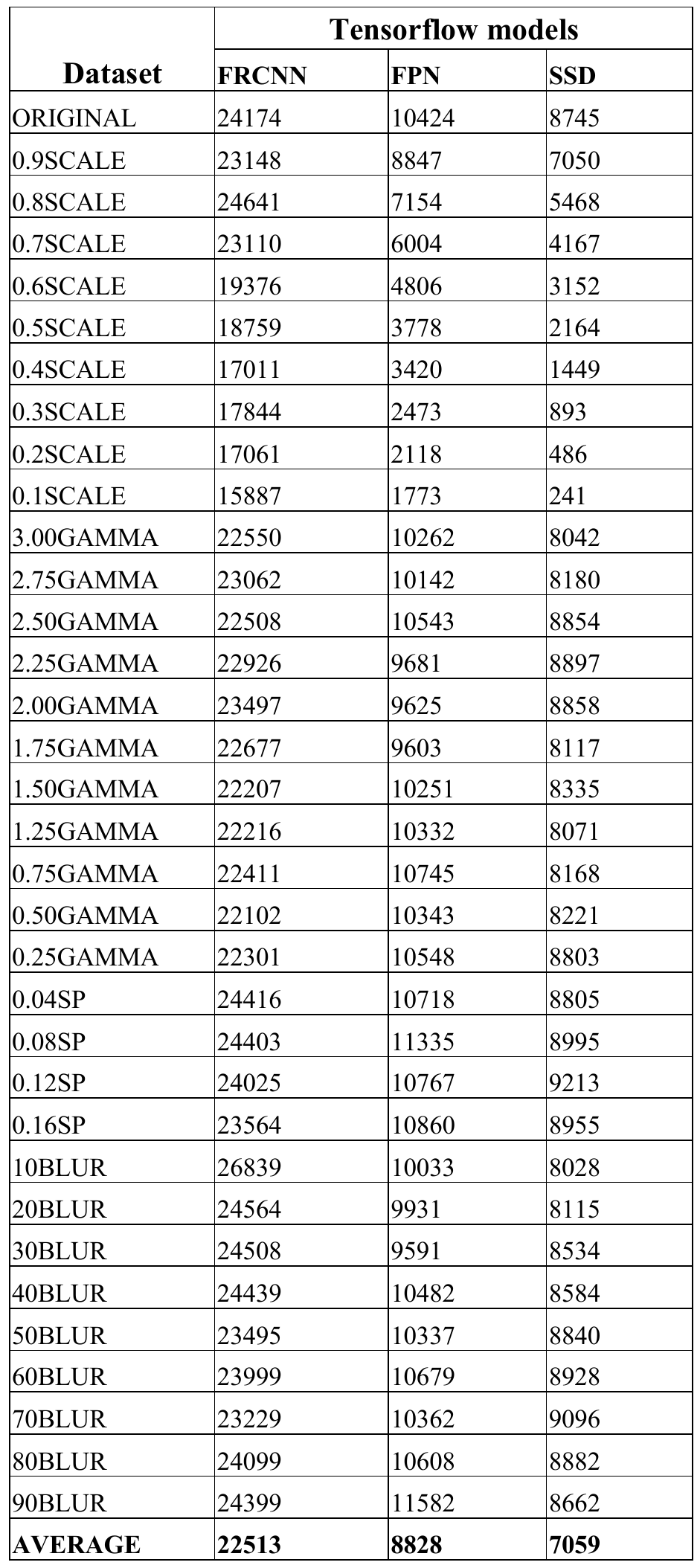}
    \caption{Average inference time per dataset 
    (in ms) }
    \label{table:acc2}
\end{table}


\newpage
With respect to our hypotheses H1-H4, the results of the  conducted study can be summarises as follows:

\begin{itemize}
    \item 
   H1 and H2 were disproved by the conducted study, see Table~\ref{table:acc1}: In the terms of accuracy, the best performing model was FPN, where FRCNN and SSD were performing slightly worse than the other models. 
    \item
    H3 was proved as correct, see Table~\ref{table:acc2} (rows presenting the results for 0.1-0.09SCALE as well as the original dataset).
    \item
    H4 was also proved as correct, see Table~\ref{table:acc2}: In average,   FRCNN was approx. 2.5 slower than FPM and approx. 3.2 times slower than SSD.  
\end{itemize}

 
The results for both Tensorflow and Anyline are also significantly better than the results of our early investigation study conducted for the domain of meter reading recognition, where we analysed Google Cloud Vision (GCV) and Amazon Web Services (AWS)  Rekognition using the same data sets. The average accuracy values for GCV and AWS Rekognition were just $29\%$ and $36\%$ respectively, see~\cite{enase2019shine}.

 For an overall comparison of the the accuracy scores over all datasets, see
  also Table~\ref{table:acc1}. 
 
On this basis, we also performed interference analysis for Tensorflow models, see Table~\ref{table:acc2}. The tests were run in the background of a laptop with an Intel i7-6700HQ CPU. However, the inference time could improve significantly (more than 10-fold) if a server-based CPU is used for the same experiment, see \cite{huang2017speed}.


\section{\uppercase{Conclusions}}
\label{sec:conclusions}

We presented the results of a research project, which goal was
 to provide an alternative, semi-automated, method for the current system to update the meter reading data, collected from  non-smart utility meters.  

Our early investigation study on the recognition accuracy of Google Cloud Vision  and AWS Rekognition applied for recognition in utility meter readings, demonstrated very low average accuracy values ($29\%$ and $36\%$, respectively). For this reasons,  
we conducted a further study to analyse two other computer vision technologies, applied for recognition in utility meter readings:
\begin{itemize}
    \item 
    an open-source Tensorflow technique (FRCNN, FPN, and SSD models), and
    \item 
    a commercial solution Anyline.
\end{itemize}
The study demonstrated that Tensorflow provides significantly better results for our application domain ($92.35\%$ for the FPN model), in comparison to Anyline, as well as to Google Cloud Vision  and AWS  Rekognition.

  This research project was conducted under the initiative 
\emph{Research embedded in teaching}, see \cite{young2021project,spichkova2019industry,simic2016enhancing,spichkova2017autonomous}.
This initiative was proposed at the RMIT University (Melbourne, Australia) within the Software Engineering projects (SEPs) conducted in collaboration with industrial partners.  
The aim of this initiative is to encourage students' curiosity for Software Engineering and Computer Science research. To reach this aim we include research components as bonus tasks in the final year projects (on both undergraduate and postgraduate levels), which typically focus on software and system development. 
Few weeks long research projects have been sponsored by industrial partners, who collaborated with the students and academic advisers through the final year projects. 
Respectively, the topics of these short research projects focus align the topics  final year projects. 
The successful results of this initiative are presented in \cite{christianto2018enhancing,clunne2017modelling,spichkova2018automated,ICECCS_SMI,enase2019shine,sun2018software,chugh2019automated,gaikwad2019voice,spichkova2019comparison,spichkova2020gosecure,george2020usage,spichkova2020vm2}. 

\emph{Future Work:}
We are going to implement a Tensorflow-based solution within the prototype we presented earlier in \cite{enase2019shine}, where the implementation was based on AWS Rekognition. The prototype includes a mobile application for automated capturing of the meter readings and managing the customer's account details, and  a Web application for management customers' accounts, details on the electricity and gas meters, etc.

\section*{{Acknowledgements}}

We would like to thank Shine Solutions and Energy Australia for sponsoring this project under the research grant RE-03615.

\bibliographystyle{plain}

\end{document}